# Form Follows Function: Recursive Stem Model

Navid Hakimi
navid.hakimi@outlook.com
github.com/navidivan/rsm

*Recursive reasoning models such as Hierarchical Reasoning Model (HRM) and Tiny Recursive Model (TRM) show that small, weight-shared networks can solve compute-heavy and NP puzzles by iteratively refining latent states, but their training typically relies on deep supervision and/or long unrolls that increase wall-clock cost and can bias the model toward greedy intermediate behavior. We introduce Recursive Stem Model (RSM), a recursive reasoning approach that keeps the TRM-style backbone while changing the training contract so the network learns a stable, depth-agnostic transition operator. RSM fully detaches the hidden-state history during training, treats early iterations as detached "warm-up" steps, and applies loss only at the final step. We further grow the outer recursion depth H and inner compute depth L independently and use a stochastic outer-transition scheme (stochastic depth over H) to mitigate instability when increasing depth. This yields two key capabilities: (i) >20× faster training than TRM while improving accuracy (≈5× reduction in error rate), and (ii) test-time scaling where inference can run for arbitrarily many refinement steps ( $\sim 20{,}000\ H_{test} \gg 20\ H_{train}$ ), enabling additional "thinking" without retraining. On Sudoku-Extreme, RSM reaches 97.5% exact accuracy with test-time compute (within ~1 hour of training on a single A100), and on Maze-Hard (30×30) it reaches ~80% exact accuracy in ~40 minutes using attention-based instantiation. Finally, because RSM implements an iterative settling process, convergence behavior provides a simple, architecture-native reliability signal: non-settling trajectories warn that the model has not reached a viable solution and can be a guard against hallucination, while stable fixed points can be paired with domain verifiers for practical correctness checks.*

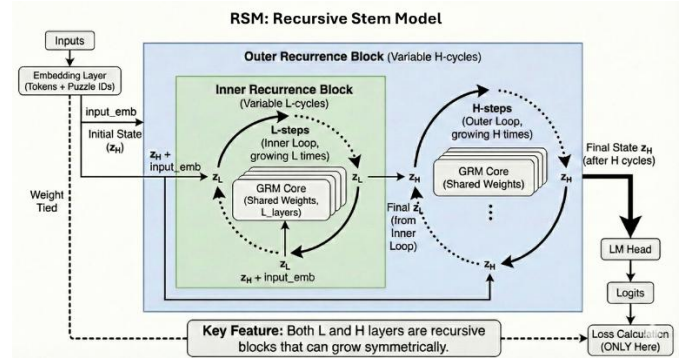

Figure 1 Recursive Stem Model Architecture

## I. Introduction

(I1) Many of the reasoning problems we care about are *verifier-rich*: the space of candidate answers is large, but correctness can often be checked quickly and deterministically (e.g., Sudoku constraints, maze-path validity, program execution), which makes them a natural target for approaches that allocate *additional computation at inference* to refine or verify solutions rather than relying on a single-shot guess. Large language models (LLMs) can be improved on such tasks via *explicit reasoning traces* and *inference-time search/selection*— for example, Chain-of-Thought prompting encourages step-by-step intermediate reasoning [1], self-consistency aggregates over multiple sampled reasoning paths [2], and "test-time compute" strategies study how to spend additional inference FLOPs (e.g., best-of-*N*, beam search, verifier-guided selection) to improve accuracy [3]. Yet these methods typically leave the underlying model class unchanged (next-token prediction), so failures often remain tied to brittleness (one wrong step can invalidate an answer) and to the cost and fragility of externalized reasoning traces—especially on benchmarks designed to probe algorithmic generalization such as ARC-style tasks [4].

(I2) A complementary direction is to build models whose *forward pass is itself an iterative improvement procedure*—a learned algorithm that can "try again" internally, not by sampling more text, but by repeatedly applying a compact transition operator. Recent looped/recursive architectures demonstrate that this can be remarkably parameter-efficient on hard puzzle domains. Hierarchical Reasoning Models (HRM) use two small networks recursing at different frequencies and train with deep supervision plus a one-step gradient approximation to avoid full backpropagation through time [5]. Tiny Recursive Models (TRM) simplify this paradigm to a single tiny network and report improved generalization on Sudoku, Maze, and ARC-style tasks with only a few million parameters [6]. Subsequent theoretical work has begun to reinterpret "deep supervision over latent improvements" as an instance of a broader improvement operator family, connecting these models to policy-improvement-like dynamics in latent space [7]. Collectively, these results suggest that recursion can serve as a *native test-time compute mechanism*—but they also surface a central training bottleneck: if improvement requires long unrolls or frequent deep supervision steps, wall-clock training and greedy intermediate steps can dominate.

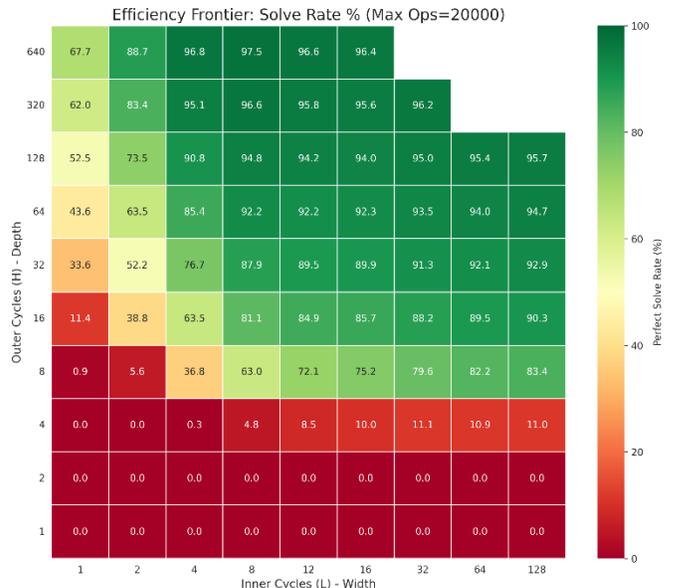

Figure 2 Perfect Solve rate for Sudoku by test time compute. The model was only trained with max depth of ~20. Previous SOTA (TRM) achieves ~87% with 12 hours of training. Notice that increasing H does not cause a drop in accuracy.

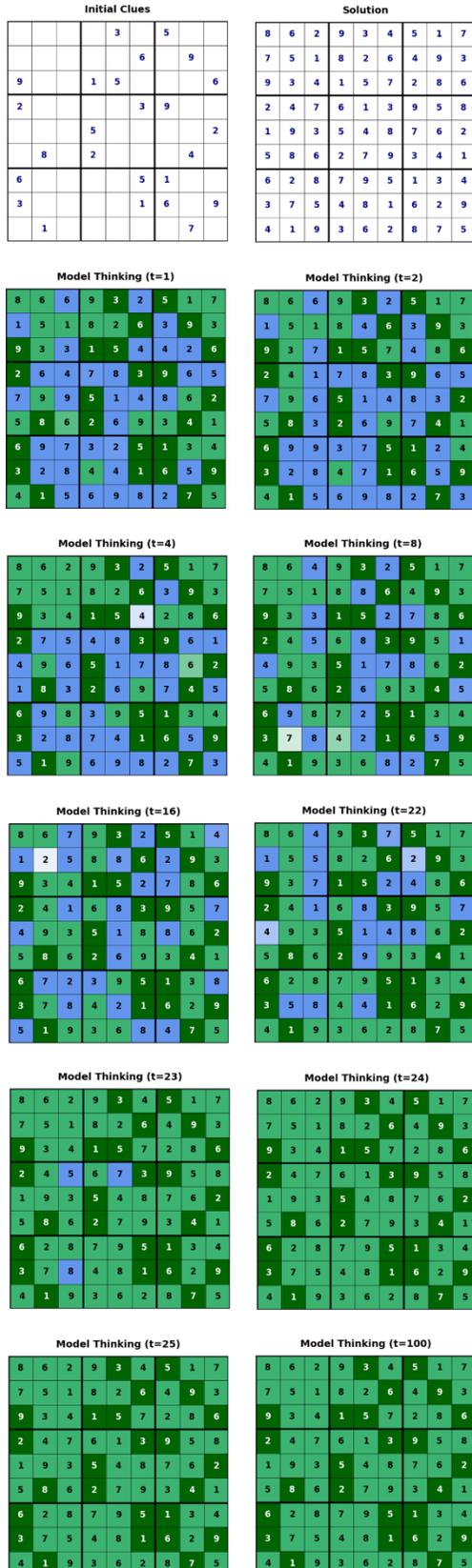

Figure 3 Solving Sudoku. Once the model finds the solution, it stops changing it!

**(I3)** In this work we introduce **Recursive Stem Model (RSM) (Figure 1)**, a recursive reasoning method that preserves the "iterate-to-improve" idea of HRM/TRM while changing the *training contract* so that the model is incentivized to learn a **stable, depth-agnostic transition function** rather than a depth-specific trajectory. Concretely, RSM (i) **fully detaches the hidden-state history** during training and applies loss only at the final step (a strict one-step/truncated gradient signal), treating earlier refinement steps as "warm-up" iterations; (ii) **grows** the effective hierarchical depths—outer depth $H$ (slow cycle) and inner depth $L$ (fast cycle)—**independently**, enabling shallow training that later scales at test time; and (iii) uses **stochastic depth** over the growing recursion to mitigate optimization instabilities when new depth is introduced [10]. This design targets two practical constraints that recur in looped reasoning models: (a) avoiding the memory/time costs and vanishing-gradient pathologies of long unrolled recurrence (the classical motivation for BPTT and its truncations) [8,9], and (b) making "more test-time compute" (**Figure 2**) a *first-class* knob by allowing inference depths far beyond training depths. Empirically (as reported in this paper), RSM achieves **97.5%** on **Sudoku-Extreme** and **80%** on **Maze-Hard** within ≈1 **hour** of training using only **2.5–5M** parameters, while reducing training time by **>20×** and error rate by **≈5×** relative to TRM under comparable settings; importantly, RSM's iterative dynamics also enable a simple *stability-based self-check* in verifier-rich domains: if the model has not settled into a stable solution-like fixed point, it should not be trusted (**Figure 3, Figure 4**). We believe that the only reason the score is not higher on Maze was the limitation of the training set considered that resulted in overfitting. Otherwise, Maze was easier for the model to master.

**(I4)** The design of RSM can be stated precisely using the language of *learned dynamical systems*: RSM parameterizes a transition operator that is repeatedly composed, and performance emerges not from a single pass but from the trajectory under iteration. This places RSM near three well-studied families of ideas—**fixed-point solvers/implicit layers**, **continuous-depth neural dynamics (Neural ODEs)**, and **iterative denoising processes**—while remaining distinct from each.

1. **Fixed-point and implicit-layer viewpoint.** Fixed-point iteration seeks $z^{\backslash *}$ such that $z^{\backslash *} = F_\theta(z^{\backslash *}; x)$; deep equilibrium models (DEQs) make this idea explicit by defining the network's "output" as a fixed point and using root-finding plus implicit differentiation to train without backpropagating through arbitrarily deep unrolls [11]. More generally, differentiable optimization layers embed argmin/feasibility operators inside networks, enabling gradients through implicit solutions of constrained problems [12]. HRM/TRM also invoke fixed-point language (and, in HRM, one-step/implicit-function-inspired gradient approximations), but do not typically solve the equilibrium exactly at training time [5,6]. **RSM's key difference** is architectural/training: it does *not* aim to compute an exact equilibrium during training, and it does *not* rely on implicit differentiation; instead, it uses a **strictly truncated** gradient signal (loss on the final step only) with **detached warm-up iterations**, forcing the learned operator to behave like a *stable improvement map* under repeated application, because the training objective repeatedly evaluates the operator after variable-depth refinement.

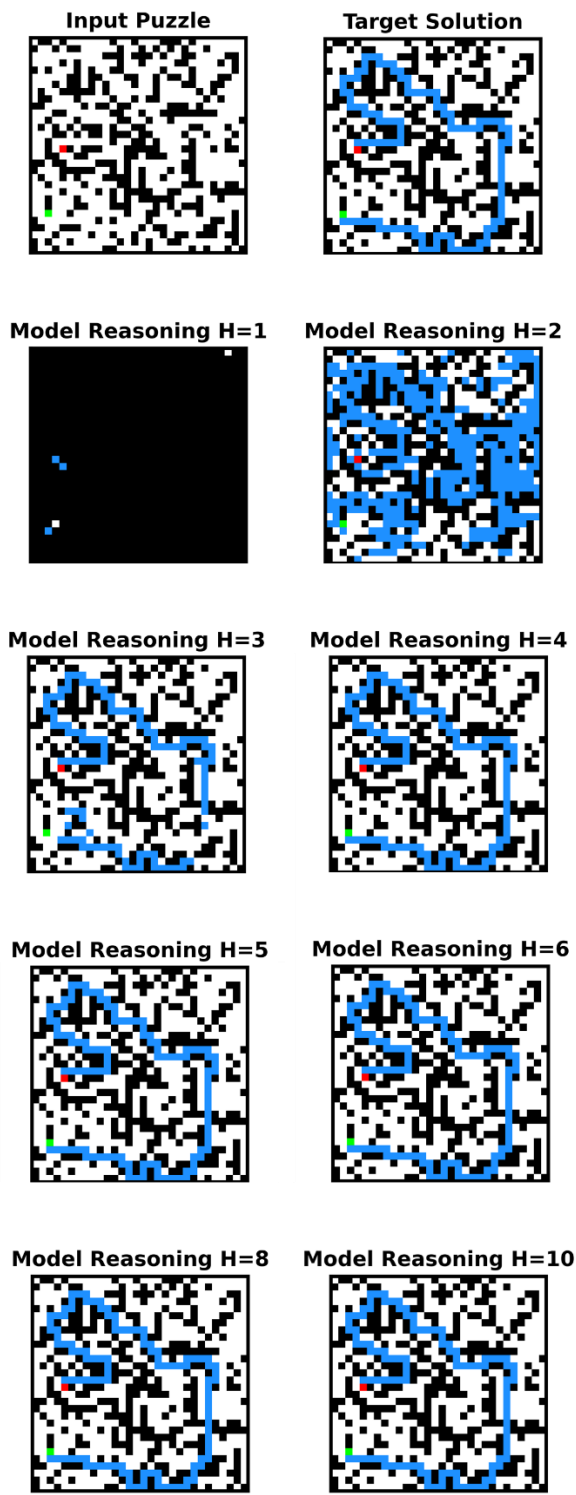

*Figure 4 Solving Maze. The strategy used by the model seems to be to first find a path at any cost and then refining it. A greedy model trained with deep supervision would likely choose a different strategy that respects color distribution.*

2. **Neural ODE / continuous-depth viewpoint.** Residual networks can be interpreted as discretizations of continuous-time dynamics, and Neural ODEs formalize learning $\dot{z} = f_\theta(z, t)$ where inference corresponds to numerical integration [13,14]. RSM is not an ODE solver, but it shares the core structural intuition: **depth corresponds to time**, and the model's effectiveness depends on whether repeated application yields a stable trajectory that approaches a task-dependent attractor. Critically, RSM is designed so that **test-time depth** can be increased far beyond training depth, echoing the way numerical solvers trade compute for accuracy by taking more steps (or smaller step sizes) at inference.

3. **Iterative denoising viewpoint.** Denoising autoencoders relate learning to map corrupted states back to data manifolds [15], and diffusion/score-based generative modeling trains a network to iteratively reverse a noise corruption process [16–19]. Sampling in these models is an explicit **refinement process** whose quality depends on the number of denoising steps; modern accelerations (e.g., DDIMs, progressive distillation, and consistency models) make this trade-off explicit by enabling fewer-step generation while preserving a multi-step "quality dial" [21–23]. RSM is not a diffusion model (there is no explicit forward noising process over the task variables), but the *architectural motif* is shared: train a compact operator that repeatedly improves an internal state, and use iteration count as a test-time compute resource. RSM's truncated-gradient training can be understood as optimizing the operator to be *robust under iteration* rather than overfitting to a specific unroll depth, which is precisely the property that makes diffusion samplers and DEQ-style solvers reusable across different step budgets.

**(I5)** We also motivate RSM using a *thermodynamics-and-boundaries* lens drawn from foundational work on life and self-organization—not as a biological claim about the brain, but as a precise analogy for why **multi-timescale recursion plus boundary-like interfaces** can be computationally powerful. Schrödinger emphasized that living systems maintain organized, low-entropy structure by sustaining a flux of energy/matter—famously framing this as feeding on "negative entropy" to resist equilibration [24]. Nonequilibrium thermodynamics and the theory of dissipative structures further formalize how ordered patterns can arise and persist under driven conditions [25,26], and contemporary statistical-physics work has explored constraints on self-replication and adaptation in energetic terms [27]. From an evolutionary perspective, multicellularity is a canonical "major transition," where cooperating units form a higher-level individual with new capabilities and new boundary conditions governing flows of nutrients, signals, and stressors [28–30]. One recurring motif in these transitions is the emergence of an *interface*—an "outer layer" that mediates interaction with the environment, effectively filtering and stabilizing the internal degrees of freedom of the collective, even when the cooperating units share the same genome. In parallel, the free-energy principle and active inference propose that the persistence of an organism can be modeled as minimizing variational free energy (a bound on surprise), with **Markov blankets** providing a formal statistical boundary that separates internal from external states while mediating sensory/active exchange [31–33]. Importantly, these ideas have been linked directly to morphogenesis: "morphogenesis as Bayesian inference" models pattern formation and control in complex biological systems using variational principles and Markov-blanket-like partitions [34]. In this paper we use these foundations to frame RSM's **two-state latent design**: $z_L$ (fast, task-conditioned computation) and $z_H$ (slower, scratchpad-like and stabilizing) act like nested interfaces that (i) separate fast computation from slower integrative state and (ii) enable "triggered" increases in effective depth, analogous to how new boundary conditions can unlock higher-level collective dynamics in development and evolution.

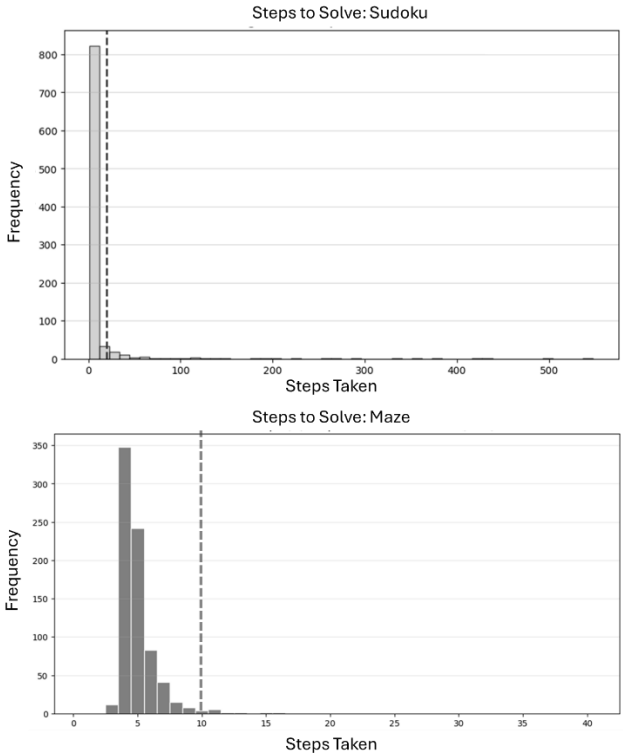

*Figure 5 Steps taken to solve. The dashed line shows maximum depth at which the models were trained at. It's worth noting that the training at that max depth was only for 5% of total steps. Also notice that the model still tries to converge early instead of dragging out the computation.*

**(I6)** Finally, RSM's $z_H/z_L$ separation suggests a concrete computational analogy to **annealing and alternating-phase optimization**, which helps situate our method among recent generative-model hybrids. Classical simulated annealing frames hard combinatorial search as stochastic exploration under a temperature schedule that gradually "cools" toward low-energy states, while **quantum annealing** replaces thermal jumps with quantum fluctuations and adiabatic evolution to traverse energy landscapes, with foundational formulations and reviews in the transverse-field Ising setting and adiabatic computation literature [37–39]. In RSM, we argue that $z_L$ can be treated as a high-entropy working state that supports broad parallel computation over partial hypotheses, while $z_H$ functions as a stabilizing scratchpad that accumulates intermediate structure and repeatedly "cools" the system toward a consistent solution—an interpretation that aligns with our emphasis on fixed-point-like settling at inference. This also connects naturally to alternating-phase learning metaphors: the wake–sleep algorithm alternates recognition and generative phases to improve latent-variable models [40], and the neuroscience literature describes complementary roles for online learning and offline consolidation/replay across hippocampal–neocortical systems [41–43]. At the level of modern sequence modeling, diffusion-based text models explicitly perform iterative refinement over an entire sequence [44,45], and newer *diffusion–transformer hybrids* move toward **blockwise or semi-autoregressive refinement**—generating and revising multiple tokens per step to interpolate between autoregressive decoding and full-sequence diffusion [46–48]. These works highlight an important constraint: pushing "generate everything in one go" tends to trade **memory and per-step compute** for fewer steps, whereas diffusion-style refinement trades **more forward passes** (and often **lower per-example sample efficiency**, since each step supervises only a subset of tokens/states) for iterative correction; indeed, recent analyses emphasize that masked/full-sequence diffusion can be harder to optimize because supervision is sparser than token-wise next-token training [48]. RSM targets a different point in this trade space: keep the model tiny and the training unroll shallow (stable gradients, fast iteration), and then reclaim the benefits of long-horizon refinement by allocating **unbounded test-time recursion** as needed—turning depth into a controllable compute budget without incurring full-history backpropagation costs during training (**Figure 5**).

## II. METHODS

This section describes **Recursive Stem Model (RSM)**: a two-state recursive reasoning model built on the TRM/HRM latent-recursion backbone, but trained to learn a **stable, depth-agnostic transition operator** by (i) **fully detaching the hidden-state history**, (ii) supervising **only the final step**, and (iii) **growing** the effective recursion depths $H$ (outer) and $L$ (inner) during training while allowing $H_{\text{test}}, L_{\text{test}}$ to scale far beyond training at inference (**Figure 6**).

```
// 1. Input Embedding
1  E_tok ← Embed(X);
2  E_puz ← Embed(P);
3  E ← Concat(E_puz, E_tok);          // Prefix puzzle embeddings

// 2. Initialize States
4  z_H ← 0;   z_L ← 0;

// 3. Nested Recurrence
5  for h ← 0 to H - 1 do
      // Detach history for warmup steps or stochastically at
      //    final step
6     if h < H - 1 or (Random() < p_detach) then
7        z_H ← StopGradient(z_H);
8        z_L ← StopGradient(z_L);

      // Inner Loop: Refine local state
9     for l ← 0 to L - 1 do
10       z_in ← z_H + E;              // Persistent Injection
11       z_L ← RSM_Block(z_L, z_in);

      // Outer Step: Update global state
12    z_H ← RSM_Block(z_H, z_L);

// 4. Prediction & Loss
13 Y_full ← LM_Head(z_H);              // Shared Weights with Input Embedding
```

Figure 6 RSM Pseudocode

### A. Problem setup and representations

We consider puzzle domains (Sudoku, Maze) cast as **sequence prediction over a discrete vocabulary**. Each training example provides:

- inputs: a token sequence $x \in \{1, ..., V\}^S$ encoding the puzzle state (e.g., grid cells and symbols),
- targets (implicit in the loss head): the desired output tokens $y \in \{1, ..., V\}^S$ (e.g., solved grid / correct per-cell labels).

RSM is **non-causal**: each output position can attend to all other positions (or be mixed by token-MLPs), matching the "solve a whole grid" nature of Sudoku and global planning dependencies in mazes. This mirrors prior TRM/HRM evaluation setups on puzzle grids

### B. RSM architecture

#### 1) Inputs and embeddings

Let $d$ denote the model hidden size. RSM first maps tokens to embeddings:

$$e(x) = \sqrt{d}\, \text{Embed}(x) \in \mathbb{R}^{B \times S \times d}.$$

Optionally, if puzzle_emb_ndim > 0, we prepend a learned sparse "puzzle embedding" $p \in \mathbb{R}^{B \times S_p \times d}$ and concatenate it to the token embeddings:

$$\tilde{e}(x) = [p; e(x)] \in \mathbb{R}^{B \times (S_p + S) \times d}.$$

In our code this is implemented by CastedSparseEmbedding and a configurable prefix length puzzle_emb_len (or computed length if puzzle_emb_len==0). The model can also apply either **RoPE** rotary position embeddings for attention-based blocks (Maze) or learned absolute positional embeddings (optional).

For Sudoku (MLP-token-mixing) we typically disable positional encodings.

Finally, all subsequent states operate on the full sequence length

$$S_{\text{tot}} = S_p + S.$$

#### 2) Two latent states: $z_H$ and $z_L$

RSM maintains two sequence-shaped latent tensors at every recursion step:

$$z_H, z_L \in \mathbb{R}^{B \times S_{\text{tot}} \times d}.$$

Intuitively (and consistent with HRM's two-timescale interpretation ), we use:

- $z_L$: a **fast** computation state (inner-cycle "working" domain),
- $z_H$: a **slow** state that accumulates and stabilizes across cycles (outer-cycle "scratchpad/answer" domain).

Initialization: in our implementation, $z_H$ and $z_L$ are initialized from fixed (persistent) buffers H_init and L_init (broadcast across batch and sequence positions). This yields:

$$z_H^{(0)} = \mathbf{1} \cdot h_0, z_L^{(0)} = \mathbf{1} \cdot \ell_0,$$

where $h_0, \ell_0 \in \mathbb{R}^d$ are fixed vectors and $\mathbf{1}$ broadcasts to shape $B \times S_{\text{tot}}$.

#### 3) The shared transition operator

RSM uses a **single shared module** (TRM-style weight sharing) applied repeatedly to both $z_L$ and $z_H$. Let $F_\theta(\cdot;\cdot)$ denote the module that:

1. adds an "injection" tensor to the current hidden state, and
2. applies $D$ stacked residual blocks (L_layers = D).

In code (RSMModule.forward):

$$F_\theta(h; u) = \text{Block}_D(\ldots \text{Block}_2(\text{Block}_1(h + u)) \ldots).$$

Each block (RSMBlock) has two configurable variants:

- **Attention variant (Maze):**
  - multi-head self-attention (non-causal) with RoPE ,
  - followed by a feed-forward MLP with **SwiGLU** gating ,
  - both wrapped in residual connections with **RMSNorm** .

- **MLP-token-mixing variant ("MLP-T", Sudoku):**
  - a token-mixing SwiGLU MLP applied across the sequence dimension (implemented by transposing to shape $B \times d \times S_{\text{tot}}$, applying an MLP on the token axis, and transposing back), conceptually similar to token-mixing in all-MLP architectures (e.g., Mixer-style token mixing),
- followed by the same channel MLP and RMSNorm residual structure.

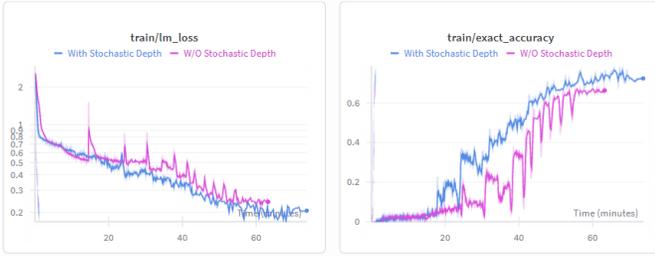

*Figure 7 The importance of stochastic depth on the stabilization of training when a new head is added.*

We tie the output projection weights to the input embedding matrix (weight tying):

$$W_{\text{out}} = W_{\text{emb}}.$$

*4) Recurrence structure: outer cycles **H** and inner cycles **L***
RSM executes a nested recurrence:
- **Outer recursion** of length $H$: $h = 1, \ldots, H$.
- Within each outer step, **inner computation** of length $L$: $\ell = 1, \ldots, L$.

At outer step $h$, RSM performs:
**Inner updates (fast, $L$ steps):**

$$z_L \leftarrow F_\theta(z_L;\ z_H + \tilde{e}(x)) \text{ repeated } L \text{ times.}$$

**Outer update (slow, 1 step):**

$$z_H \leftarrow F_\theta(z_H;\ z_L).$$

This is exactly how *rsm*.py executes the loops: at each outer cycle it runs L_cycles updates to z_L using injection z_H + input_emb, then updates z_H using injection z_L.

**Prediction.** After the final outer cycle, we decode logits from $z_H$:

$$\text{logits} = W_{\text{out}}\, z_H^{(H)} \in \mathbb{R}^{B \times S_{\text{tot}} \times V},$$

and discard any prefix positions corresponding to puzzle embeddings, returning logits for the original token positions.

### C. Training objective: terminal loss with detached warm-up

The key departure from TRM-style deep supervision is **how gradients are propagated through recurrence**.

*1) Why not full unrolling or deep supervision?*
Backpropagation Through Time (BPTT) unrolls recurrence and differentiates through the full trajectory, but is memory intensive and susceptible to vanishing/exploding gradients with long horizons. Truncated BPTT reduces cost but still couples learning to a particular unroll horizon. Deep supervision (supervising intermediate depths) can stabilize optimization and improve gradient flow in deep networks, but for recursive reasoning it can also bias the model toward **greedy intermediate behavior** (optimizing every step to look "good" rather than optimizing convergence of the iterative process). RSM instead trains the model to **arrive** at a correct fixed point, not to be correct at every intermediate step.

*2) Terminal loss*
Let $y$ be the ground-truth output tokens and let $\hat{y}^{(H)}$ be the final-step prediction from logits at outer depth $H$. RSM computes loss **only at the final step**, e.g. token-level cross entropy:

$$\mathcal{L}(\theta) = \text{CE}\big(\text{logits}^{(H)}(\theta), y\big),$$

with no auxiliary losses at intermediate outer cycles or inner cycles.

*3) Full detachment of history ("warm-up" iterations)*
During training we run multiple outer cycles, but **detach** the hidden states to prevent gradients from flowing through most of the history. Concretely, at the beginning of many outer steps we do:

$$z_H \leftarrow \text{stopgrad}(z_H),\ z_L \leftarrow \text{stopgrad}(z_L),$$

which cuts the computational graph while preserving the forward iterative refinement.

We conceptually partition outer cycles into:
- **Warm-up steps:** forward refinement steps that shape the state but carry no gradient history.
- **Active step:** the final step where the loss is computed and gradients flow through the inner and outer updates.
- **(Optional) transition step:** a single additional step immediately preceding the final step that can be included in the gradient path

This "detach-and-refine" scheme forces RSM to learn a transition operator that is useful **after arbitrary warm-up**, rather than one that relies on gradient credit assignment over a long, depth-specific trajectory.

*4) Stochastic depth over the outer transition (training $H - 1$ or not)*
When training deep recursive models, increasing $H$ can induce sudden loss spikes (distribution shift in recursion depth and in the latent state distribution). We mitigate this by a stochastic depth–style regularization applied specifically to the **outer transition step** (Figure 7).

Classical stochastic depth randomly drops residual layers during training while keeping them at test time, improving optimization and generalization in very deep nets. RSM adapts this idea to recursion depth by stochastic **graph connectivity** rather than forward compute: we always run the forward warm-up steps, but we stochastically decide whether to allow gradients to span the final two outer steps.

*5) Two-step vs one-step gradient in outer depth*

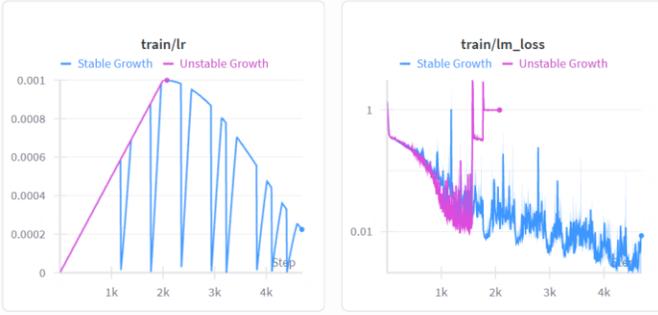

*Figure 8 The Importance of resetting learning rate for the transition periods on the training of the transformer architecture*

Let the final outer step index be $h = H$. RSM always performs the final-step updates with gradients enabled. Optionally, we also include a "transition" outer update at $h = H - 1$ in the gradient path.

- **If transition is included:** gradients flow through the outer update at $h = H - 1$ and then through the final active step at $h = H$, yielding an effective **two-step** gradient span in outer depth.
- **If transition is excluded:** the final step detaches its inputs and gradients flow only within the final step, yielding a strict **one-step** truncated gradient.

This is implemented by the include_prev_H logic in rsm.py. The hyperparameter prob_detach_prev_H controls the probability of excluding the transition (i.e., "detach previous $H$").

At each training forward pass, we sample:

$$\text{include\_prev\_H} \sim \text{Bernoulli}(1 - p_{\text{detach}}),$$

where $p_{\text{detach}}$ = prob_detach_prev_H.

- If include_prev_H = False, we detach at the active step and do not create a gradient connection to the penultimate step.
- If include_prev_H = True, we keep the active step connected to the penultimate outer update.

In our released training recipes, we typically set $p_{\text{detach}}$ close to 1 (e.g., 0.99 for Sudoku), making training predominantly one-step while still occasionally exposing the optimizer to a slightly longer horizon. We enforce a small but important constraint: during training we clamp the requested outer depth to $H \geq 2$ (see H = max(2, H_req) if training), ensuring there is always at least one warm-up step even in early curriculum phases. During inference we allow $H = 1$.

### D. Training objective: terminal loss with detached warm-up

*1) Growing recursion depths: independent curricula for $H$ and $L$*

A central design goal is to **separate training stability from inference compute** by growing recursion depth gradually. We implement an explicit curriculum over both:

- $H$: outer recursion depth (H_cycles),
- $L$: inner refinement depth (L_cycles).

*2) Milestone-based schedules*

Let training progress be measured in percent:

$$p = 100 \cdot \frac{\text{step}}{\text{total\_steps}}.$$

We start from base values $H_0$ and $L_0$ (from config). Given milestone lists $\{(m_i, \Delta H_i)\}$ and $\{(n_j, \Delta L_j)\}$, we set:

$$H(p) = H_0 + \sum_i \mathbf{1}[p \geq m_i]\Delta H_i, \quad L(p)$$
$$= L_0 + \sum_j \mathbf{1}[p \geq n_j]\Delta L_j.$$

This is exactly the logic in get_target_H() and get_target_L() in pretrain.py. At each training step we compute active_H, active_L, override $H$ at forward time (override_H_cycles) and update the model config for $L$ (model.cfg.L_cycles = active_L). This design makes "effective depth" a tunable resource: training can remain shallow early (fast and stable), while later stages increase compute as needed.

*3) Transition stabilization*

Depth changes can destabilize optimization. Our training loop includes several stabilizers:

- **Gradient clipping:** we clip global gradient norm (default 1.0) to reduce exploding-gradient events.
- **Learning-rate warmup and cosine schedule:** we use a standard warmup + cosine decay schedule. (**Figure 8**)
- **Optional transition LR warmup:** after a depth transition we optionally ramp the effective LR from 0 to 1 over transition_lr_warmup_steps (see lr_transition_mult()), which reduces shock after adding compute.
- **EMA of parameters:** optionally maintain an exponential moving average of model weights for evaluation stability (Polyak-style averaging).
- **Optimizer-state scaling (optional):** when depth grows, we can rescale momentum buffers by a factor (optimizer_reset_scale) to control how much the optimizer "forgets" pre-transition statistics.

the request

### E. Inference-time scaling and convergence diagnostics

At inference we run the same recurrence but may choose much larger $H_{\text{test}}$ and/or $L_{\text{test}}$ than used in training:

$$H_{\text{test}} \gg H_{\text{train}}, L_{\text{test}} \gg L_{\text{train}}.$$

To analyze RSM as an iterative solver, we record the **rollout** $\{\hat{y}^{(h)}\}_{h=1}^{H_{\text{test}}}$ by decoding after each outer cycle. This supports:

- **Steps-to-solve:** the first outer step $h^{\backslash *}$ at which the decoded solution satisfies the verifier (and, in our analysis plots, remains stable thereafter).
- **Fixed-point / settling checks:** measure whether $\hat{y}^{(h)}$ stops changing for consecutive steps, used as a practical diagnostic for convergence behavior in verifier-rich domains.

These diagnostics correspond to the rollout visualizations and steps-to-solve plots reported in our experiments.

III. DISCUSSION

Tiny Recursive Models (TRM) and Hierarchical Reasoning Models (HRM) revived a simple but powerful idea: **reasoning depth can come from recursion**, not from huge parameter counts. In both lines, a small transition function is applied repeatedly to refine an internal latent state until it encodes an answer, yielding strong puzzle performance under small-data regimes. RSM takes this paradigm one step further by treating recursion explicitly as **operator learning**: rather than training a model to be correct at a fixed unroll depth, we train it to implement a **stable state update rule** that can be applied *for arbitrary durations at test time*. This framing makes RSM naturally comparable to (i) fixed-point and implicit-depth models such as Deep Equilibrium Models (DEQs), where solutions are defined as equilibria of a learned map, and (ii) continuous-depth viewpoints such as Neural ODEs, where computation is "more steps" of a learned dynamics. Our empirical takeaway is pragmatic: by **training shallow but deploying deep**, RSM can achieve high accuracy quickly and then leverage additional inference steps as "test-time compute" for hard instances.

*1) Why terminal-only training + full detachment changes the learned computation*

A central decision in RSM is to **completely detach** the recurrent history during training and apply supervision **only at the final step** (i.e., no deep supervision across intermediate steps). This differs sharply from classic Backpropagation Through Time (BPTT), which differentiates through an unrolled trajectory, and from truncated BPTT, which differentiates through a fixed window. It also differs from deep supervision schemes that attach auxiliary losses to intermediate layers/steps to stabilize gradients.

Conceptually, detachment turns most early steps into **warm-up iterations**: they shape the state distribution seen by the final step, but they do not receive credit assignment through time. The effect is to push the model away from "memorize the trajectory at depth $H$" and toward "learn a map that works from many nearby states," i.e., a **state transition operator** that can be re-applied. This is closely aligned with the intuition behind fixed-point models (DEQs) where one cares about convergence to an equilibrium rather than the path taken to get there.

A useful way to think about this is: *deep supervision rewards being right early*, while terminal-only supervision rewards *ending up right*. In domains like Sudoku and Maze—where intermediate states may be partially inconsistent yet still useful for search—this difference matters, because forcing early "correct-looking" states can bias the model toward greedy refinement rather than stable convergence.

*2) Independent growth of $H$ and $L$: separating "how we learn" from "how long we think"*

RSM's second design choice is to **grow** the two compute axes independently:

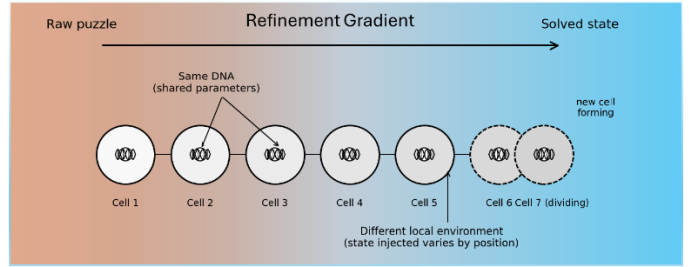

Figure 9 *Recursive Stem Model RSM growth as a 1D organism along an interior–exterior gradient: proliferation occurs at the deepest, most stable end, pushing older cells outward toward the environment—analogous to stem-cell niches in hair follicles and the intestine epithelium lining.*

- $H$: the number of **outer cycles** (global refinement iterations)
- $L$: the number of **inner cycles** (local compute per outer step)

Training begins with shallow effective depth and gradually increases $H$ and $L$ via a curriculum. The practical value is that it decouples:
**(a) optimization stability during training** from **(b) inference-time compute available at test**.

This decoupling has strong precedents across ML systems that "train short / test deep." Stochastic depth, for example, trains very deep residual networks by randomly dropping layers during training but using full depth at test time, explicitly to improve optimization and generalization. In generative modeling, *iterative refinement* families—masked iterative decoding and diffusion—also embody the idea that "more steps" yields better quality, and that compute can be shifted between training and sampling. Mask-Predict (iterative masked decoding) refines an entire sequence by repeatedly re-masking low-confidence tokens, and MaskGIT does the same for image tokens, showing large speedups by predicting many tokens per iteration and refining over multiple rounds.

Recent diffusion language models formalize this multi-step refinement for text, but they expose a recurring trade-off: **global refinement steps are expensive because each step touches the whole sequence**, and training–inference alignment becomes nontrivial. Diffusion-LM and DiffuSeq exemplify iterative denoising for text generation, while SSD-LM explicitly adopts a **semi-autoregressive** strategy that generates **blocks** of tokens iteratively (multi-token per step) to balance quality and speed. More recently, block diffusion work has emphasized that the *granularity of supervision* should match the *granularity of decoding* (e.g., "blockwise SFT"), reinforcing the broader point that step structure matters.

RSM occupies a different point in this landscape: it is not a generative diffusion model, but it **shares the compute philosophy**: (i) learn an operator that refines state, (ii) let inference steps scale, and (iii) manage the memory/compute trade-off explicitly. Detaching history makes this especially clean: **training memory is roughly constant in recursion depth**, and thus very deep test-time rollouts become feasible without re-training or storing long computation graphs.

*3) Stochastic outer-transition training: a recursion-native analogue of stochastic depth*

Our third core mechanism is a recursion-native adaptation of stochastic depth: at training time, we **randomly decide** whether gradients will flow through the penultimate outer step ($H-1$) or not, while always training the final step $H$. This is implemented as a stochastic rule that sometimes excludes the "transition" step from the gradient path, effectively creating a mixture of:

- **1-step outer credit assignment** (strictly final-step training), and
- **2-step outer credit assignment** (final step + one transition step).

This choice directly addresses a phenomenon visible in our loss curves: **when we grow recursion depth, loss can spike** (a distribution shift in the latent-state dynamics). The stochastic transition rule smooths this by preventing the optimizer from being forced into a single brittle horizon and by exposing it to a controlled variety of short credit-assignment paths.

The analogy to stochastic depth is tight: in stochastic depth, training sees "shallower effective depth" while test uses full depth; in RSM, training sees "shorter effective temporal depth" while inference can use very long rollouts.

*4) Interpreting $z_L$ and $z_H$: compute vs commit, warm-up vs cool-down*

You clarified an important conceptual point, and it is worth stating crisply:

- $z_L$ **is the *computation domain*.** It is where "thinking happens": a high-dimensional, distributed, *superposed* internal representation that can explore and transform hypotheses without immediately committing to a crisp answer.
- $z_H$ **is the *commit/cool-down domain*.** It is where the system "writes things down": it consolidates what was discovered in $z_L$, stabilizes it, and makes it available as the next iteration's context and ultimately as the readout state.

This "compute vs commit" framing is often more faithful than a purely "hierarchical timescale" metaphor. Mechanistically, RSM's update rules already encode a strong asymmetry: the inner computation updates $z_L$ using injections derived from $z_H$ (and the puzzle input), and then the outer update updates $z_H$ using $z_L$. That is, the model alternates between a **high-plasticity transform space** and a **stabilized memory-like state**, over and over.

A useful *thermodynamic* metaphor is annealing: you can interpret $z_L$ as a "higher temperature" computation phase (rapid, exploratory, high-entropy transformations), and $z_H$ as a "cooling" phase (projection and stabilization). In classical simulated annealing, stochastic exploration is gradually cooled to settle in low-energy configurations; quantum annealing replaces thermal fluctuations with quantum fluctuations and a schedule that reduces them over time. We are **not** claiming RSM implements quantum processes; the value is the structural analogy: alternating updates can be seen as repeated cycles of exploration and consolidation that encourage convergence to a stable configuration (a fixed point).

This perspective also resonates with broader "life as a fight against entropy" narratives. Schrödinger's *What is Life?* famously discusses how living systems maintain order (negentropy) despite the second law, framing life as sustained organization far from equilibrium. Prigogine and Nicolis' work on dissipative structures similarly treats self-organization as an emergent property of nonequilibrium systems driven by flows and constraints. Within this lens, RSM's $z_L \rightarrow z_H$ alternation is a computational caricature of a deeper principle: **systems that remain stable while performing work often require separate regimes for transformation and for stabilization**.

Finally, the "commit" language naturally invites a memory analogy. The wake–sleep algorithm alternates phases driven by recognition vs generative connections, as a biologically motivated way to train deep generative models. And modern neuroscience theories of memory consolidation emphasize that sleep can support reactivation and reorganization of memories, changing their representations and integrating them with existing structure. Complementary Learning Systems theory explicitly argues for two learning systems with different timescales—fast episodic and slow integrative—to avoid catastrophic interference while still enabling rapid acquisition. Again: RSM is not a brain model, but the analogy helps articulate why separating "fast compute" from "stable commit" can produce both efficient learning and stable long-horizon inference.

*5) A multicellular analogy: shared "DNA," different function via position and context*

A second conceptual lens you raised is biological and—importantly—**not** about genetic differences:

In multicellular organisms, cells share the same DNA, yet they do different things based on where they are situated and what signals they receive.

RSM has a strikingly similar structural property: every recursion step applies **the same parameters** (shared transition "DNA"), yet each step can behave differently because it is "prompted" by a different internal context—specifically, a different $z_H$ and $z_L$ shaped by prior computation and by proximity to the puzzle/solution boundary.

This maps naturally onto the idea of **positional information**: cells interpret spatially varying signals (often described as gradients) to produce context-appropriate behavior. Wolpert's classic formulation of positional information frames development as cells acquiring a coordinate-like identity and responding accordingly. If we treat the RSM rollout as a **1D organism** (**Figure 9**) stretched along a gradient between "raw puzzle" and "solved state," then each outer step is like a cell at a different "position" in that gradient. The computation is identical in rules (shared parameters) but different in effect because the local "environment"—here, the state injected via $z_H$—differs.

This analogy also ties to the broader theme of **major evolutionary transitions**, where collections of units become integrated into higher-level agents. Maynard Smith and Szathmáry emphasize multicellularity as a transition in how information and control are organized across parts of a system. In RSM, recursion plus shared weights yields something similar in miniature: instead of independent modules, we have repeated applications of the same module whose collective behavior produces emergent problem-solving capacity.

Mark Levin's work on morphogenesis and bioelectric control emphasizes that "cognition-like" dynamics can exist at scales below brains, with tissues navigating attractor landscapes in "morphospace." Kuchling, Friston, Georgiev, and Levin explicitly connect morphogenesis to a variational / inference perspective, casting pattern formation and control in terms of minimizing variational free energy. From the Friston side, Markov blankets formalize how boundaries allow systems to maintain identity and regulate exchange with their environment. If we squint, $z_H$ plays a boundary-like role: it is the stabilized interface between the raw input and the internal exploratory computations in $z_L$, and it is the object that persists through many iterations and is ultimately "presented" as the system's output. This is, again, a metaphor—but it is a useful one for articulating why RSM can be both tiny and powerful: **the "organism" is not in encoded complexity in weights; it is emergent complexity in repeated interaction.**

*6) Fixed points, "knowing when it's wrong," and what convergence can (and cannot) guarantee*

One of RSM's most practically interesting behaviors is that long rollouts often show **progressive refinement** followed by **stability**: after a certain number of iterations, the model's decoded solution stops changing and remains fixed. This invites a natural stopping rule: treat convergence to a fixed point as evidence of "I'm done thinking."

Fixed-point framing is standard in DEQs (the model *is* defined by its equilibrium), and it is also implicit in iterative denoising generative models where the objective is to reach a clean sample after enough refinement steps.

However, there is an important scientific caveat: **convergence is not correctness** in general. A system can converge to a wrong fixed point (a stable but invalid attractor), or oscillate between a small set of states. What RSM gives you is a *new diagnostic signal* that purely autoregressive decoders often lack: you can distinguish "still changing" from "settled," which can be combined with domain verifiers (Sudoku validity checks, maze path correctness) to create robust stopping and confidence measures. In verifier-rich settings, the conjunction "settled + passes verifier" is a strong success certificate; "not settled" is a strong warning sign of incomplete reasoning.

*7) Limitations and forward directions*

**(0) Parameter Sweeps and Ablation Study.** This paper in its current form is presented as a concept paper and focuses on introducing novel architecture and co-evolution of architecture with training. Its scope does not include finding best parameters and combinations of factors, nor does it focus on systematic ablation studies, due to budget and resources constraints. Total budget for all the experiments and runs was 50$ (around 0.5$ per run) on Google Colab. We only tried a limited number of training schedules, learning rates, batch sizes, stochastic factors, etc. as well as base model complexities, and tuning was more artistic than systematic.

**(i) Convergence is not guaranteed.** RSM is trained to behave like a stable operator, but without explicit contraction constraints it may still oscillate or settle to spurious attractors. DEQ-style techniques—explicit equilibrium solving and implicit differentiation—offer one direction for making the equilibrium viewpoint more principled, though at different engineering trade-offs.

**(ii) Training–inference alignment remains a deep theme.** Our curricula and stochastic transition training are explicit attempts to align what the model learns with how we will use it. The diffusion-LM literature has independently rediscovered how critical "step structure alignment" is (e.g., blockwise fine-tuning to match blockwise decoding). A longer-term goal is to develop general principles for step-aligned supervision in iterative models—recursive, masked-refinement, or diffusion.

**(iii) The compute–memory frontier is still open.** Modern diffusion and masked-refinement models show many ways to trade "number of steps" against "tokens per step," including semi-autoregressive block generation (SSD-LM) and masked diffusion recipes that bring perplexity closer to AR baselines. RSM contributes a complementary approach: keep memory cheap during training via detachment, then spend inference compute where it matters.

**(iv) Optimal Growth.** We tested a limited set of scheduled in this work. The questions remains open: what's the optimal rate of growth? Can we determine apriority conditions for stability and conditions of growth based on the information theoretic and/or thermodynamics frameworks? Can we determine the timing of growth based on the behavior of gradiants and/or the internal activations? What can we learn from biology, the prepreation of cells and the environmental signals for structuring and timing of mitosis?

**(v) Beyond puzzles.** Sudoku and Maze are ideal for studying iterative reasoning because correctness is verifiable and rollouts are interpretable. Extending RSM -style recursion to open-ended generation (where multiple valid outputs exist) will likely require integrating explicit verification, constraint satisfaction, or preference-based signals—otherwise fixed points may encode "consistent" but undesired solutions.

## IV. CODE AVAILABILITY

The code accompanying this paper can be found at github.com/navidivan/rsm
.